\documentclass[conference]{IEEEtran}
\pdftrailerid{}
\usepackage{cite}
\usepackage{amsmath}
\usepackage{booktabs}
\usepackage{array}
\usepackage{graphicx}
\usepackage{float}
\usepackage{capt-of}
\newcolumntype{L}[1]{>{\raggedright\arraybackslash}p{#1}}
\usepackage{xcolor}
\usepackage{tikz}
\usepackage{pgfplots}
\pgfplotsset{compat=1.18}
\usetikzlibrary{arrows.meta,positioning}
\usepackage[hidelinks]{hyperref}

\newif\ifanon
\anonfalse
\ifanon
  \newcommand{\sys}{\textsc{Bound}}
\else
  \newcommand{\sys}{PlugRL}
  \hypersetup{pdftitle={Detecting Silent Faults at the Environment-Trainer
    Boundary in Reinforcement Learning}, pdfauthor={Zuo Gou, Chenhao Lu}}
\fi

\newcommand{\msg}[1]{\texttt{#1}}

\begin{document}

\title{Detecting Silent Faults at the Environment--Trainer Boundary
in Reinforcement Learning}

\ifanon
\author{\IEEEauthorblockN{Anonymous Authors}}
\else
\author{\IEEEauthorblockN{Zuo Gou\textsuperscript{*} and
Chenhao Lu\textsuperscript{*}}
\IEEEauthorblockA{Tsinghua University, Beijing, China\\
\{gouz23,\,luch21\}@mails.tsinghua.edu.cn\\
\textsuperscript{*}Equal contribution}}
\fi

\expandafter\ifx\csname iflatexml\endcsname\relax
  \expandafter\newif\csname iflatexml\endcsname
  \latexmlfalse
\fi

\newcommand{\teasercaption}{A value changed at the environment--trainer
boundary raises no error, the policy still learns, and only checks of the run
itself notice. (a) Schematic: one observation is changed on its way to the
trainer; a record on each side of the boundary, compared step by step (the
step ledger), disagrees at that step. (b)
Pendulum, SB3's PPO through \sys{} (E54): the mean return of each batch of 16
episodes in the trainer's log. Blue, the median and range of the
boundary-fault runs at seeds 0--2 with one value changed, the 57 of 63 that
logged all 32 batches (arrow: the median step at which the value changed);
grey, the median and range of ten fault-free seeds; dashed, the gain
threshold, 500 above the runs' median start. (c) The 63 runs, one cell each
and in the same order in every row, grouped by the field the fault changed.
Filled: the check caught the run (step ledger: E55).}

\iflatexml\else
\makeatletter
\let\IEEEaftertitletext\@IEEESAVECMDIEEEaftertitletext
\makeatother
\IEEEaftertitletext{%
\vspace{-0.6\baselineskip}%
\noindent\begin{minipage}{\textwidth}
\centering
\includegraphics[width=\textwidth]{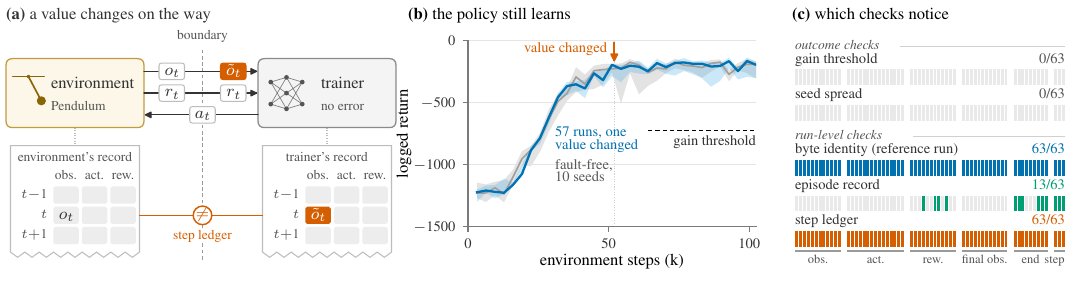}
\captionof{figure}{\teasercaption}
\end{minipage}
\vspace{0.4\baselineskip}}
\fi

\maketitle
\iflatexml
\begin{figure*}[t]
\centering
\includegraphics[width=\textwidth]{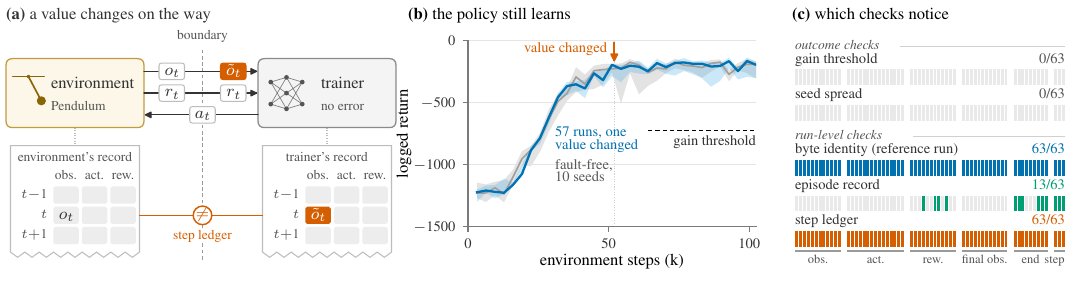}
\caption{\teasercaption}
\label{fig:teaser}
\end{figure*}
\else
{\makeatletter\edef\@currentlabel{\arabic{figure}}\label{fig:teaser}\makeatother}
\fi

\begin{abstract}
Robot reinforcement learning increasingly runs environments outside the
trainer: in another process, on another machine, in another language. Many
faults at this boundary raise no error. We make the boundary a wire
protocol and ask what notices a fault in it.

On Pendulum, HalfCheetah, Hopper and a 7-joint arm (Pusher), we injected up
to 23 faults into what crosses the wire and 2 into the trainer's log, from
one value to every value. Unless every value was wrong, a learning-curve
threshold caught at most one run in ten, and a comparison with ten
fault-free seeds at most 21\%. A clean evaluation of all 1,060 final
policies agreed with the curves. Of the 728 boundary-fault runs that passed
both curve checks, 723 scored no more than three standard deviations below
fault-free seeds, yet nearly all ended on other weights and, for reward
faults, logged other returns.

Checks of the run itself, not of its outcome, saw every fault. A step
ledger kept on both sides and compared step by step caught and located
every fault that crossed, on Pendulum, Hopper and Pusher on one machine and
on Pendulum across two, with no fault-free run and, on Pendulum, a median
3\% more training time. An episode record kept at the environment caught
the faults in the trainer's log. We list such faults found in practice, in
Gymnasium 1.3 and our own code and, reported by others, in five multi-agent
libraries, and recommend a step ledger and episode counts kept at the
environment.
\end{abstract}

\section{Introduction}

An RL run has a boundary between the code that steps the environment and
the \emph{trainer}, which collects experience and updates the policy: the
agent--environment interface~\cite{sutton2018}, drawn in software. In many
stacks nobody designs it; it falls where the trainer puts it. A Gymnasium environment~\cite{towers2024gymnasium} is an
object in the trainer's process. RLlib~\cite{liang2018rllib},
SRL~\cite{mei2024srl} and RLinf~\cite{yu2025rlinf} run environments in
workers that install the framework and join its cluster. Robot learning
moves the boundary further: to simulators in other languages, machines
with no GPU stack, and robots reaching a policy over the
network~\cite{black2024pi0,cadene2024lerobot}.

Many faults at this boundary are silent: they raise no error or
warning~\cite{tambon2024silent}. Under its \texttt{SAME\_STEP}
autoreset mode, Gymnasium~1.3's vector episode statistics drop the first
reward of every episode after an environment's first, and count its length
one short, until 1.4 (Sec.~\ref{sec:wild}). Training goes on. For environments on
another machine, our own evidence that the boundary was right was
that the policy still learned.

We test that evidence. \sys{} makes the boundary an explicit wire protocol
(Sec.~\ref{sec:protocol}). The environment side sends observations, applies
the actions it gets back, and feeds back rewards and episode ends. A
written specification and conformance checkers say what each side must do,
and lockstep bridges let Stable-Baselines3 (SB3)~\cite{raffin2021sb3},
CleanRL~\cite{huang2022cleanrl} and RLinf train through it. On one machine
a run through it is deterministic, so we can compare a run with an injected
fault against the same seed without it. We call a difference between what
the environment took and gave and what the trainer sent and received a
\emph{boundary fault}. Our method is mutation testing as
RLMutation~\cite{tambon2023rlmutation} applies it to deep RL, and the checks
that work are the ones production ML uses against training--serving
skew~\cite{breck2019datavalidation}; what we add is measuring both at the
environment--trainer boundary. We find:

\begin{itemize}
\item \textbf{Neither a learning curve nor a clean evaluation shows most
  boundary faults} (Sec.~\ref{sec:faults}, Fig.~\ref{fig:teaser}). We
  injected stale, zeroed, swapped or rounded observations, actions and
  rewards, wrong episode ends and skipped steps. On Pendulum, the gain
  threshold we had relied on caught 1 of 21 faults at doses up to 10\%
  of values, and 9 when every value was wrong. On a clean environment, 723
  of the 728 boundary-fault runs that passed the curve checks scored at
  most three standard deviations below fault-free runs.
\item \textbf{A step ledger catches and locates every boundary fault, as it
  must, and cheaply} (Sec.~\ref{sec:faults}). A boundary fault is a
  difference between the two sides' records, so comparing them must catch
  it. Wherever we kept one (585 runs, three tasks) it did; we measured that
  it needs no fault-free run or determinism, works across machines, names
  the step, environment and field, and on Pendulum costs a median 3\% of
  training time. An episode record kept at the
  environment catches faults in the trainer's log.
\item \textbf{Such faults occur in practice} (Sec.~\ref{sec:wild}): in
  Gymnasium 1.3 and our own code, and, reported by others, in five
  multi-agent libraries.
\end{itemize}
A silent fault, then, mostly changes the run rather than the policy: its weights
and, for reward and log faults, the numbers it reports. The record of the
experiment stops describing the experiment.

Every experiment follows a protocol written before its registered runs,
most after a pilot. Changes to it go in a dated amendment. Captions,
tables and appendices name each result's experiment (E43, \ldots), whose
protocol, logs and findings are public with the code, specification and
checkers: \url{https://github.com/PlugRL}.

\section{The Boundary as a Protocol}
\label{sec:protocol}

\textbf{Messages.} A connection opens with the server's metadata (version,
chunk length $H$, offered features). The client then repeats three messages
(Fig.~\ref{fig:checks}), msgpack-encoded with NumPy arrays over a
WebSocket. An \msg{infer} carries the observations of $n$ environments,
their indices and their step ids; an \msg{action} reply carries a chunk of
$H$ actions; after the chunk, a \msg{feedback} carries the observation it
ended on, the summed reward, the termination and truncation flags, and an
info map. The client resets its own environments: the feedback that ends an
episode carries the terminal observation, and the next infer the reset one.
A connection closes with \emph{stop} (the run is over) or \emph{resync}
(reconnect at once).

\textbf{What the protocol does about faults.}
\begin{itemize}
\item \emph{One convention, written down.} The specification fixes one
  episode-end convention where Gymnasium has several (Sec.~\ref{sec:wild}),
  and states every MUST clause. A checker for each side tests against
  them, with active probes.
\item \emph{Structural faults are loud.} The bridge rejects a message with
  the wrong shape, keys or environments and closes the connection. It never
  fills in a missing step: losing a client stops the run, and a client too
  many is told to stop.
\item \emph{Runs are deterministic on one machine.} The bridge is a lockstep
  server: each trainer step sends every environment its action, and returns
  once every client has fed back. So two runs through it can be compared
  byte for byte.
\item \emph{Each side can keep its records.} A wrapper at the env client
  can keep an episode record and half a step ledger; the bridge keeps a bridge
  log of the rewards it received.
\end{itemize}
Adapters present an episode end as RLinf, Gymnasium (\texttt{SAME\_STEP})
and SB3 expect it, so these trainers use the protocol through the
environment interface they already have.

\section{What Notices a Fault}
\label{sec:faults}

\begin{figure}[t]
\centering
\includegraphics[width=\columnwidth]{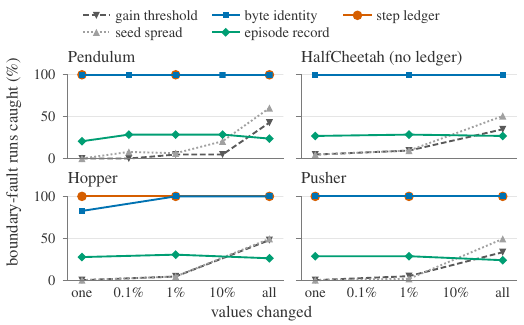}
\caption{Only the step ledger catches every boundary fault at every dose
it ran; byte identity misses only Hopper's 12 final observations, which
training never read. The share of boundary-fault runs each check caught, seeds
0--2: 21 faults, 63 runs per dose (23 and 69 on Hopper). Pendulum and
HalfCheetah E54, Hopper E56, Pusher E58; step ledger E55, E56, E58. The
ledger was not run on HalfCheetah.}
\label{fig:faults}
\end{figure}

\textbf{The faults.} SB3's PPO trained Pendulum,
HalfCheetah, Hopper and Pusher, a 7-joint arm that pushes an object to a
goal, through the bridge (Appendix~\ref{app:e50}), while the env client changed one
field of what it sent, in one way. The observation it sent, the action it
applied, the reward, or the observation an episode ended on was replaced by
zeros, by the previous step's value or by the next environment's, or
rounded through float16; an action could also be clipped to half its range.
An episode end was invented, or a time limit left out or sent as a
termination; on Hopper, which ends an episode when it falls, a termination
was also left out or sent as a time limit. Or a step was skipped and its
feedback filled in. Two log faults sat in the trainer's own episode log, of the kind we found
in Gymnasium (Sec.~\ref{sec:wild}): each episode's first reward left out,
or its length counted one short. Each fault changed one value, 1\% or every
value it could reach (on Pendulum also 0.1\% and 10\%), the same values at
every seed, seeds 0--2.

\textbf{The checks} (Fig.~\ref{fig:checks} in the appendix). A run's fault was caught by
the \emph{gain threshold} if its \emph{gain}, the mean logged return at the
end of training minus that at the start, fell below a bar fixed before the
runs (on Pendulum, 500); by
\emph{seed spread} if its final return lay more than three standard
deviations from the mean of ten fault-free seeds; by \emph{byte identity}
if its final weights differed from the fault-free run's at the same seed;
by the \emph{episode record} if the trainer's log disagreed with the env
client's record, kept before any fault, in count, length or return; and by
the \emph{step ledger} if the two sides' records of any step differed.

\textbf{Learning curves miss them} (Fig.~\ref{fig:faults}). At doses up to
10\% of values, the gain threshold caught at most one run in ten on all four
environments. With every value wrong it caught 43\%, 35\%, 48\% and 33\%,
and mostly faults that broke learning or emptied the log. The seed-spread
check raised few false alarms: 0, 1, 0 and 0 of ten fault-free seeds, each
judged against the other nine. Yet at doses below every value it caught at
most 21\%. The spread of seeds explains it. One changed value moved a
boundary-fault run's final return by a median of 15, 85, 197 and 0.7; changing the seed moved it
about as much or more, by standard deviations of 11, 146, 259 and 1.9.

\begin{figure}[t]
\centering
\includegraphics[width=\columnwidth]{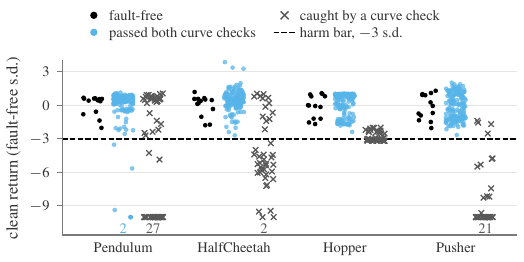}
\caption{Nearly all policies behind faults that pass both curve checks
stay above the harm bar. Each final policy's clean return, the mean of 50
deterministic episodes on a fresh environment (E57, E58), in standard
deviations of ten fault-free seeds. Boundary-fault runs at seeds 0--2, every
dose; black, fault-free runs. Points below $-10$ sit on the floor, counted.}
\label{fig:clean}
\end{figure}

\textbf{So does a clean evaluation.} We ran each of the 1,060
final policies for 50 deterministic episodes on a fresh environment
(Fig.~\ref{fig:clean}). Of the runs that passed both curve checks, 5 of 255
on Pendulum and none of 473 on the other three scored more than three
standard deviations below ten fault-free seeds; the other 106 runs that did
had all failed a curve check. With every reward zeroed, the log reads 0, the best possible
return on Pendulum and on Pusher, while the policies score $-1196$ to
$-1639$ and $-99$ to $-129$.

\textbf{A step ledger sees every boundary fault.} A wrapper
around each environment, outside the env client's code, recorded each
step: the observation it had given, the action it was handed, and what it
returned. A second wrapper, around the vector environment SB3 trained on,
recorded what it sent and received. A boundary fault
is a difference between the two, so the comparison must catch it; the test
is whether it does so in practice, and at what cost. Compared step by
step, the two halves caught all 189 boundary-fault runs on Pendulum, 207 on
Hopper and 189 on Pusher, each at its first changed value and field, and
found no difference in the 27 runs on each without one. The two added a
median 3\% to Pendulum's training time ($-5$ to $+9$\%, each run's
\texttt{learn\_s} against the same run without them) and left all 216 runs' weights unchanged.
With the env client on a laptop and SB3
on a workstation, they caught and located all 21 faults at one value, and
found no difference in the fault-free run. That run ended on other weights
than on the workstation: byte identity cannot judge across machines.

\textbf{The episode record sees the books.} It caught every log fault, and
every boundary fault that changed a reward beyond its tolerance, an episode
end or some steps. It missed faults in observations, final observations and
actions, ends sent as the wrong kind, and every step skipped: none changes
the episodes it compares. The bridge log, kept on the trainer's
side of the wire, caught only the log faults. A record that is to catch a
client's faults has to be kept at the environment.

\textbf{Byte identity sees what training reads.} On one machine the
weights changed in all 504 boundary-fault runs on Pendulum and HalfCheetah,
and in none of 48 runs with a 1\,ms delay, a float64 wire or a log fault.
On Pusher, whose episodes all end at the time limit, they changed in 189 of
189. On Hopper they changed in 195 of 207, and in none of its 24 such runs.
In the other 12,
one final observation changed, at a fall. SB3 bootstraps from it only at a
time limit, so training never read it, and the weights stayed
byte-identical. The fault was real, and the step ledger caught all 12; a
trainer that reads that value would have learned from it. The two faults
the bridge rejects, rows out of order and a reward missing, raised within
10\,s.

\textbf{Why it matters when the policy learns.} What these faults do, no
curve check sees. They change the record: in 888 of 900 runs the final
weights differ from the fault-free run's at the same seed, and a reward or
log fault changes the returns the trainer logs, as Gymnasium's
\texttt{SAME\_STEP} statistics did (Sec.~\ref{sec:wild}). They can cost the
policy: 5 of the 255 Pendulum runs that passed both curve checks, under
five different faults and doses, scored more than three standard
deviations below fault-free seeds, though 13 fault-free runs cannot tell
that from a rare bad seed. In one arm of a comparison, a reward or
log fault would read as a difference between
methods~\cite{henderson2018matters}. Whether a fault reaches the weights
depends on how the trainer bootstraps, as Hopper's final observations show
under SB3; on harder tasks
such faults can change what is learned, as time-limit handling
does~\cite{pardo2018time}.

\section{Faults Found in Practice}
\label{sec:wild}\label{sec:silent}

The faults of Sec.~\ref{sec:faults} are not hypothetical
(Table~\ref{tab:silent}). We found some while building and checking \sys{},
many by reconciliation or a controlled comparison, and none by an error.
Others were reported in multi-agent libraries, where agents that act in
turn or leave early add conventions to get wrong. Every boundary fault
among them is a kind we injected. Faults in a run's lifecycle and cost,
which fall outside the boundary, are in Appendix~\ref{app:outside}.

\begin{table}[t]
\centering
\caption{Faults that raised no error, found in practice.}
\label{tab:silent}
\footnotesize
\setlength{\tabcolsep}{3pt}
\begin{tabular}{@{}L{0.17\columnwidth}L{0.31\columnwidth}L{0.46\columnwidth}@{}}
\toprule
kind & where & what was wrong \\
\midrule
reward & RLlib~\cite{ray31156} & an agent with no last observation got
  reward 0, not the environment's \\
 & Tianshou 0.4.2~\cite{tianshou399} & rewards earned while the opponent moved
  were dropped \\
action & PettingZoo 1.22~\cite{pettingzoo939} & one-dimensional
  continuous actions were cast to integers \\
final obs. & SuperSuit~\cite{supersuit232} & the reset's info
  overwrote the terminal observation \\
episode end & MAPPO's SMACv2 wrapper~\cite{onpolicy86} & dead agents' padded
  steps were trained on as real \\
log & Gymnasium 1.3 (E51) & under \texttt{SAME\_STEP}, every episode after
  an env's first lost its first reward and a step (fixed in
  1.4~\cite{gymnasium1624}) \\
 & CleanRL~\cite{huang2022cleanrl} (E51) & on Gymnasium 1.3, not the 0.29
  it pins, no episode was logged \\
 & RLlib 2.53~\cite{ray59860} & a shared module logged one agent's
  return, not the sum \\
 & our specification (E44) & for a client written from it, the server
  logged a return of 0 \\
\bottomrule
\end{tabular}
\end{table}

\section{Moving the Boundary}
\label{sec:moving}

We applied the checks to the protocol itself (Table~\ref{tab:moving},
Appendix~\ref{app:moving}). On one machine, SB3 trained 16 Pendulum,
HalfCheetah or Atari environments in its own process and behind the
bridge: all nine pairs ended on byte-identical weights and logged identical
episodes, and all 12 Pendulum runs behind a proxy that held
every byte 0--25\,ms each way ended on the in-process weights. The env client waits for each action,
so latency decides when a step happens, not what it produces. The bridge
added 0.4--0.6\,ms a step for 16 state vectors and 2.6--3.0\,ms for 16
frames. Across machines a seed no longer reproduces: two MuJoCo builds
differ in the last bit at the first step. Every seed passed its
gain threshold across machines, with env clients on a Windows laptop with
no torch, Ray or RLinf, and on one machine, in a C++ program of
under 900 lines and under RLinf, SB3 and CleanRL. Sec.~\ref{sec:faults} shows that threshold to be weak; the step
ledger's one fault-free run across machines, clean at every step, is the
stronger evidence.

\section{The Price of the Boundary}
\label{sec:price}

On one machine an exchange costs 0.1\,ms with states only and under 1\,ms
at 588\,KiB. Between two machines it costs a latency $s$ plus $k$ copies of
the observation's $P$ bytes over the bandwidth $B$, $t = s +
kP/B$~\cite{hockney1994communication}, which predicted the measured times
within 3--10\%. The negotiated \texttt{reuse-feedback-obs} feature
sends each observation once instead of twice: at 588\,KiB the round trip
fell from 49.8 to 27.1\,ms, with byte-identical weights
(Appendix~\ref{app:price}). A 184\,KiB camera observation crosses our Wi-Fi
path in 11.6\,ms, 12\% of the 100\,ms we measured for one action chunk from
a full-size $\pi_{0.5}$~\cite{intelligence2025pi05} in openpi~\cite{openpi2025}.
A full-size $\pi_{0.5}$ with LIBERO weights took one learning step through
the protocol on LIBERO, which says nothing of how well RL tunes it.

\section{Conventions We Recommend}
\label{sec:conventions}

Do not take a learning curve alone as evidence that the boundary is right
(Sec.~\ref{sec:faults}). Instead:
\begin{itemize}
\item \textbf{Keep a step ledger.} Record what each environment took and
  gave, in a wrapper around it, and what the trainer sent and received, and
  compare the two step by step. Wherever we kept one (Pendulum, Hopper,
  Pusher), it caught and located every boundary fault we injected, on one
  machine and on Pendulum across two, for a median 3\% of training time on
  Pendulum.
\item \textbf{Count episodes at the environment.} Keep the episode record
  next to the environment, not on the trainer's side of the wire, and check
  the trainer's log against it.
\item \textbf{On one machine, show that a change beneath the policy leaves
  the weights alone}, byte for byte: a change of placement, transport,
  latency, scheduler or protocol version
  (Secs.~\ref{sec:moving}--\ref{sec:price}, Table~\ref{tab:outside}). If it
  cannot, say why.
\item \textbf{Report the boundary}: where the environments ran, how they
  reached the trainer, the bytes per step and the time per step. State the
  episode-end convention, distinguish \emph{stop} from \emph{resync}, and
  never fill in a missing step.
\end{itemize}

\section{Related Work}

\textbf{What beneath the algorithm changes results.} Implementation details
change PPO's results~\cite{engstrom2020implementation,huang2022details},
time limits change what is learned~\cite{pardo2018time}, reporting
practice changes what can be
concluded~\cite{henderson2018matters,agarwal2021precipice}, and a one-bit
change moves a run as far as a new seed~\cite{summers2021nondeterminism}.
We add the environment--trainer boundary to that list. Our faults are
silent~\cite{tambon2024silent}, of kinds found in DRL
programs~\cite{nikanjam2022faults}. Our design is mutation
testing~\cite{demillo1978hints} as DeepCrime~\cite{humbatova2021deepcrime}
and RLMutation~\cite{tambon2023rlmutation} apply it, and RLMutation found
that comparing mean rewards misses mutants because training is stochastic.
The step ledger is an end-to-end check~\cite{saltzer1984endtoend} that bridges
what each side perceives~\cite{huang2017gray}, TFX's training--serving
skew check~\cite{breck2019datavalidation} applied to the boundary.

\textbf{Environments behind a channel.} RL-Glue~\cite{tanner2009rlglue}
made the interface language-independent. ML-Agents~\cite{juliani2018unity},
Godot RL Agents~\cite{beeching2021godot} and RLlib~\cite{liang2018rllib}
reach environments outside the trainer; SEED~RL~\cite{espeholt2020seed},
SRL~\cite{mei2024srl} and RLinf~\cite{yu2025rlinf} keep them inside the
framework. dm\_env\_rpc~\cite{dmenvrpc} has a compliance
suite, the closest precedent for our checkers, but no trainer side.
Gymnasium's \texttt{check\_env}~\cite{farama2023envchecker} and EnvPool's
tests~\cite{weng2022envpool} check an environment against its API or a
reference, before training; range and NaN checks would pass a stale
or rounded value. openpi~\cite{openpi2025} serves policies for inference
only; LeRobot's HIL-SERL~\cite{cadene2024lerobot} runs the policy beside the
robot. None checks a run's record against one kept at the
environment.

\section{Limitations}

We chose the faults, to cover every field but info and the ways
Table~\ref{tab:silent} shows it going wrong, and injected them into four
simulated, state-based environments with dense rewards, under one trainer
(SB3's PPO) at three seeds. That the faults the curves miss leave the
policy intact may not hold for sparse rewards, long horizons, replay
buffers, images or real robots. The step ledger needs a wrapper around the
environment, and sees nothing that goes wrong inside it; a robot reached
only through its own client can be wrapped no closer than that client. We
compared ledgers after each run (150\,MB a run on Hopper); checking a
digest of each step as the run goes is future work. The same people wrote
the specification and the C++ client. We do not study environments that
keep moving while the env client waits for an action, as a real robot does.

\ifanon\else
\section*{Acknowledgment}
We thank Zilin Kang, who helped build PlugRL, and the reviewers for their
comments.
\fi

\bibliographystyle{IEEEtran}
\bibliography{refs}

\appendices
\raggedbottom

\section{What Each Check Caught, Fault by Fault}

\begin{figure}[H]
\centering
\includegraphics[width=\columnwidth]{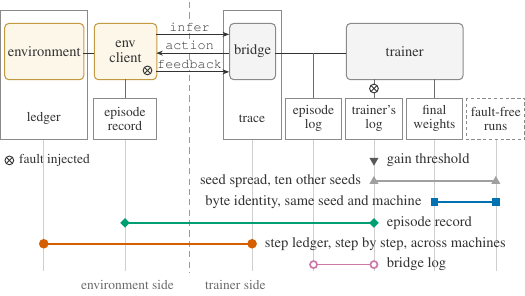}
\caption{The boundary and the checks. Filled boxes run code; outlines are
records; the dashed outline is other runs. Three messages cross the
boundary. Faults ($\otimes$) were injected in the env client, after its
episode record, and in the trainer's log. Each row marks what a check
compares; only the episode record and the step ledger reach across the
boundary.}
\label{fig:checks}
\end{figure}

Fig.~\ref{fig:heatmap} breaks Fig.~\ref{fig:faults}'s Pendulum panel down
by fault and adds the log faults and the bridge log.

\section{Faults Outside the Boundary}
\label{app:outside}

Two more kinds of fault raised no error while we built and checked \sys{}:
in a run's lifecycle and in its cost (Table~\ref{tab:outside}).

\begin{table}[H]
\centering
\caption{Faults outside the boundary that raised no error.}
\label{tab:outside}
\footnotesize
\setlength{\tabcolsep}{3pt}
\begin{tabular}{@{}L{0.17\columnwidth}L{0.31\columnwidth}L{0.46\columnwidth}@{}}
\toprule
kind & where & what was wrong \\
\midrule
lifecycle & RLinf (E48) & environments never closed; outside clients
  reconnected 3,190 times in 25 minutes \\
 & our bridge (E48) & a surplus client was told to resync, not stop: 2
  million reconnects in 10 minutes \\
cost & our scheduler (E47) & a 0.1\,ms sleep that the event loop rounds to
  1\,ms made collection 1.87$\times$ slower; the fix left the weights
  byte-identical \\
\bottomrule
\end{tabular}
\end{table}

\section{Moving the Boundary in Full}
\label{app:moving}

E43 found the two machines' MuJoCo builds differ by $8.9\times10^{-16}$ at
the first step. The bridge added 0.37--0.58\,ms per vector step of 16
state vectors (E50) and 2.60--2.99\,ms of 16 frames (E53).
On one machine, E49 also compared every trainer's log with the bridge log,
and they agreed; kept on the trainer's side, the bridge log rules out log
faults, not boundary faults. RLinf learns Pendulum late: its budget
was raised to 409,600 steps after its pilot, before the registered runs.

\begin{table}[H]
\centering
\caption{Moving the boundary. Above the line, byte identity with the
in-process run; below it, the pre-registered gain threshold (seeds that
pass / seeds run).}
\label{tab:moving}
\footnotesize
\setlength{\tabcolsep}{2pt}
\begin{tabular}{@{}llll@{}}
\toprule
exp. & trainer & environments ran & result \\
\midrule
E50 & SB3, PPO & in-process vs.\ bridge & same weights 6/6 \\
E53 & SB3, CNN & Atari frames, as E50 & same weights 3/3 \\
E52 & SB3, PPO & proxy, 0--25\,ms each way & same weights 12/12 \\
\midrule
E43 & \sys{}, FPO~\cite{mcallister2025fpo} & same machine & passes 6/6 \\
    & (HalfCheetah~\cite{todorov2012mujoco}) & laptop, Wi-Fi + Tailscale & passes 6/6 \\
E44 & \sys{}, PPO~\cite{schulman2017ppo} & Python env client & passes 3/3 \\
    & (Pendulum)         & C++, no third-party library & passes 3/3 \\
E48 & RLinf, PPO         & same machine & passes 3/3 \\
    & (HalfCheetah)      & laptop, Tailscale & passes 3/3 \\
E49 & RLinf, SB3 (PPO) & C++ Pendulum & passes 6/6 \\
    & SB3, CleanRL (PPO) & HalfCheetah  & passes 6/6 \\
\bottomrule
\end{tabular}
\end{table}

\section{The Price of Crossing}
\label{app:price}

E43 checked $t = s + kP/B$ on its two machines, and E46 measured what
\texttt{reuse-feedback-obs} saves (Fig.~\ref{fig:price}). The 100\,ms of
Sec.~\ref{sec:price} is E10's: the median, over three launches, of one call
to openpi's \texttt{sample\_actions} (five denoising steps) for a
full-size $\pi_{0.5}$ with random weights, which change a forward's output
but not its cost, on one GPU. The $\pi_{0.5}$ learning step is E11's.

\begin{figure}[H]
\centering
\includegraphics[width=\columnwidth]{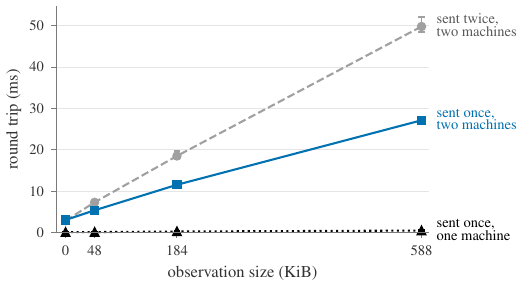}
\caption{Sending each observation once nearly halves a large exchange
between machines. Median round trip per exchange, five runs of 1,000
exchanges each (bars: their range), over one Wi-Fi and Tailscale path at
23.7\,MB/s and on one machine (E46). Without \texttt{reuse-feedback-obs}
each observation is sent twice; with it, negotiated when the connection
opens, once.
``States only'' is plotted at 0.}
\label{fig:price}
\end{figure}

\begin{figure*}[t]
\centering
\includegraphics[width=\textwidth]{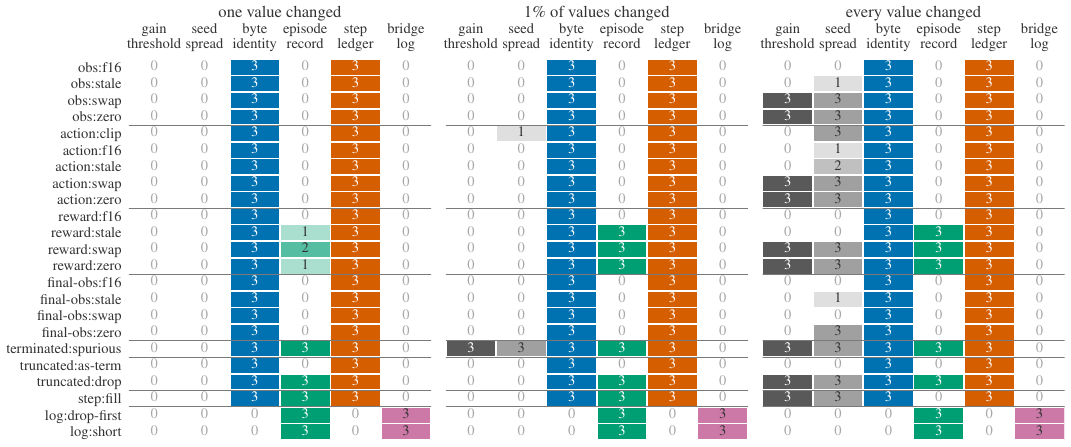}
\caption{What each check caught on Pendulum, fault by fault (E54; step
ledger E55): how many of the three seeds 0--2 it caught, with one value,
1\% and every value changed. Rows are grouped by the field the fault
changed; the last two rows are log faults.}
\label{fig:heatmap}
\end{figure*}

\section{What the Environment Side Installs}
Table~\ref{tab:footprint} measures, in fresh environments, what each
system's environment side installs to exchange data with its trainer (E45).
A thin environment side is not unique: dm\_env\_rpc's~\cite{dmenvrpc} and
openpi's client~\cite{openpi2025} are thinner than our full env client. Of the systems
here that train through their channel, only \sys{}'s environment side
needs no torch.

\begin{table}[H]
\centering
\caption{Environment side of each system (E45): RLlib 2.58, LeRobot 0.6.1,
dm\_env\_rpc 1.1.7. Sizes exclude the interpreter.}
\label{tab:footprint}
\footnotesize
\setlength{\tabcolsep}{3pt}
\begin{tabular}{@{}lccrc@{}}
\toprule
system & trains & packages & size & torch \\
\midrule
\sys{}, C++ client & yes & 0 & 86\,KB & no \\
\sys{}, minimal Python & yes & 2 & 2.6\,MB & no \\
\sys{}, full env client & yes & 31 & 222\,MB & no \\
RLlib external env~\cite{liang2018rllib} & yes & 64 & 6.1\,GB & CUDA \\
LeRobot HIL-SERL~\cite{cadene2024lerobot} & yes & 144 & 6.1\,GB & CUDA \\
dm\_env\_rpc~\cite{dmenvrpc} & no trainer & 12 & 81\,MB & no \\
openpi client~\cite{openpi2025} & inference only & 9 & 90\,MB & no \\
\bottomrule
\end{tabular}
\end{table}

\section{E50 in Full}
\label{app:e50}

SB3~2.9.0 PPO, 16 environments, 16$\times$256 frames per rollout, one torch
thread, on a 24-core Linux workstation. Pendulum uses rl-zoo's
settings~\cite{raffin2020zoo} for 102,400 steps; HalfCheetah uses SB3's
defaults for 409,600 steps. In-process means DummyVecEnv. Bridge means the
same 16 environments in another process, on one connection to the bridge's
SB3 adapter, a \texttt{VecEnv}. A vector step is one step of all 16
(Table~\ref{tab:e50}).

\begin{table}[t]
\centering
\caption{E50 per seed: the final weights' SHA-256 (first eight digits) and
the wall time, in-process and through the bridge.}
\label{tab:e50}
\footnotesize
\setlength{\tabcolsep}{3pt}
\begin{tabular}{@{}lccrrr@{}}
\toprule
env & seed & weights (SHA-256) & in-process & bridge & per vector step \\
\midrule
Pendulum    & 0 & \texttt{ec5b8226} both & 12.7\,s & 16.3\,s & +0.57\,ms \\
            & 1 & \texttt{062e92e3} both & 12.7\,s & 16.5\,s & +0.58\,ms \\
            & 2 & \texttt{3a8d968f} both & 12.7\,s & 16.5\,s & +0.58\,ms \\
HalfCheetah & 0 & \texttt{8261c03a} both & 64.0\,s & 76.0\,s & +0.47\,ms \\
            & 1 & \texttt{86564694} both & 66.7\,s & 76.5\,s & +0.38\,ms \\
            & 2 & \texttt{65cc8843} both & 67.0\,s & 76.5\,s & +0.37\,ms \\
\bottomrule
\end{tabular}
\end{table}

Every pair also logged identical episodes. The bridge log agreed with SB3's
\texttt{monitor.csv} in every bridge run.

\section{E51 in Full}

Every counter ran, with Gymnasium 1.3.0, on an environment whose step $t$
pays reward $t$ and which terminates at $t=5$. Every episode therefore has
return 15 and length 5. There were 2 environments with 3 episodes each
(Table~\ref{tab:e51}). Gymnasium had already fixed its vector counter on
its main branch~\cite{gymnasium1624}; the fix was released in 1.4.0, after
these runs.

\begin{table}[H]
\centering
\caption{E51: what each episode counter logged, as (return, length).}
\label{tab:e51}
\footnotesize
\setlength{\tabcolsep}{3pt}
\begin{tabular}{@{}L{0.5\columnwidth}lL{0.27\columnwidth}@{}}
\toprule
counter & autoreset & logged \\
\midrule
Gymnasium vector \texttt{RecordEpisodeStatistics} & \texttt{NEXT\_STEP} & $6\times(15,5)$ \\
\quad the same & \texttt{SAME\_STEP} & $2\times(15,5)$, then $4\times(14,4)$ \\
single-env \texttt{RecordEpisodeStatistics} per sub-env & both & $6\times(15,5)$ \\
CleanRL's logging loop, verbatim & \texttt{NEXT\_STEP} & nothing \\
\quad the same & \texttt{SAME\_STEP} & \texttt{TypeError} \\
SB3 \texttt{VecMonitor}; SB3 \texttt{Monitor} & SB3's & $6\times(15,5)$ \\
bridge, SB3 and Gymnasium adapters, and the bridge log & \texttt{SAME\_STEP} & $6\times(15,5)$ \\
our env client's counter, as its runner uses it & its own & $6\times(15,5)$ \\
\quad the same, with no explicit resets & \texttt{NEXT\_STEP} & $(15,5)$, $(30,11)$, $(45,17)$ \\
\bottomrule
\end{tabular}
\end{table}

The last row breaks that counter's unwritten contract: only an
explicit \texttt{reset}, which its runner always calls, zeroes it.

\end{document}